\documentclass[runningheads]{llncs}
\usepackage[T1]{fontenc}
\usepackage{xcolor}
\definecolor{lightred}{RGB}{255, 102, 102}

\usepackage[normalem]{ulem} % \sout を使えるように

\usepackage{multirow}
\usepackage{subcaption}
\usepackage{threeparttable}
\usepackage{caption}
\usepackage{wrapfig}
\usepackage{hyperref}
\usepackage{amsmath}  % 数学記号のための拡張機能
\usepackage{amsfonts} % \mathbb と他の数学フォントのため
\usepackage{amssymb}  % 追加の数学記号のため
\usepackage{algorithm}
\usepackage{booktabs} % For formal tables
\usepackage{array}    % For centering in table cells
\usepackage{graphicx}
\usepackage{xcolor}
\usepackage{algpseudocode}
\algnewcommand\algorithmicinput{\textbf{Input:}}
\algnewcommand\algorithmicoutput{\textbf{Output:}}
\algnewcommand\Input{\item[\algorithmicinput]}
\algnewcommand\Output{\item[\algorithmicoutput]}
\begin{document}
\title{Contrastive Learning-Enhanced Trajectory Matching for Small-Scale Dataset Distillation}
\titlerunning{Dataset Distillation with Contrastive Learning}
\author{Wenmin Li\inst{1} \and
Shunsuke Sakai\inst{1} \and
Tatsuhito Hasegawa\inst{1}}%
\authorrunning{F. Wenmin Li et al.}

\institute{$^1$Graduate School of Engineering, University of Fukui\\
  3-9-1 Bunkyo, Fukui City, Fukui 910-8507, Japan, \\
  \email{mf240815@g.u-fukui.ac.jp}  
}
\maketitle              % typeset the header of the contribution
\begin{abstract}
Deploying machine learning models in resource-constrained environments, such as edge devices or rapid prototyping scenarios, increasingly demands distillation of large datasets into significantly smaller yet informative synthetic datasets. Current dataset distillation techniques, particularly Trajectory Matching methods, optimize synthetic data so that the model’s training trajectory on synthetic samples mirrors that on real data. While demonstrating efficacy on medium-scale synthetic datasets, these methods fail to adequately preserve semantic richness under extreme sample scarcity.
To address this limitation, we propose a novel dataset distillation method integrating contrastive learning during image synthesis. By explicitly maximizing instance-level feature discrimination, our approach produces more informative and diverse synthetic samples, even when dataset sizes are significantly constrained.
Experimental results demonstrate that incorporating contrastive learning substantially enhances the performance of models trained on very small-scale synthetic datasets. This integration not only guides more effective feature representation but also significantly improves the visual fidelity of the synthesized images. Experimental results demonstrate that our method achieves notable performance improvements over existing distillation techniques, especially in scenarios with extremely limited synthetic data.

\end{abstract}

\section{Introduction}
\label{sec:Introduction}
Model distillation \cite{journals/corr/HintonVD15}, proposed in 2015, aims to transfer the "dark knowledge" embedded in a pre-trained teacher model into a smaller student model, enabling the student to achieve comparable performance with a significantly lighter architecture. Data distillation \cite{Radosavovic_2018_CVPR}, introduced in 2017, improves semi-supervised learning by leveraging large amounts of unlabeled data along with labeled data. Inspired by both approaches, dataset distillation \cite{wang2020datasetdistillation} was proposed in 2018 with the goal of compressing a large-scale dataset into a much smaller synthetic dataset that maintains comparable performance when used for model training. This technique accelerates training, reduces storage and computational costs, and holds great potential for deploying models in resource-constrained environments \cite{Kosugi_2024}.

\begin{figure}[h]
    \centering
    \includegraphics[width=\linewidth]{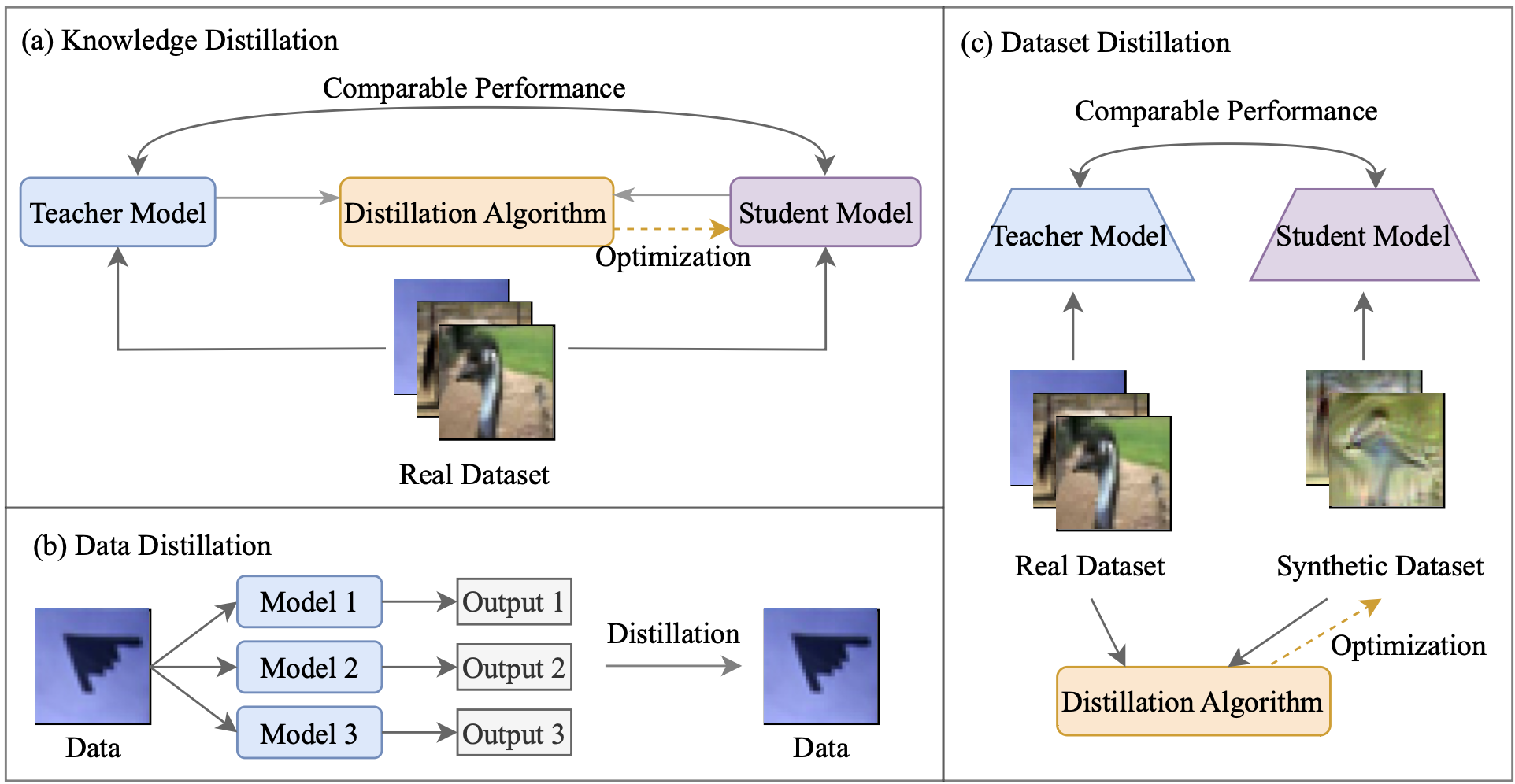}
    \caption{(a) The teacher model transfers knowledege to the student model through distillation algorithm, achieving performance parity.(b) Multiple models produce outputs from the same data, and these outputs are aggregated to refine the data itself. The distilled data retains the same input appearance but is enriched with aggregated label or prediction information, making it more informative for training.(c) The synthetic dataset is optimized to reproduce the training trajectory or generalization behavior of real data, enabling the student model to achieve comparable performance to the teacher model.}
    \label{fig:introduction}
\end{figure}

Datset distillation has seen rapid progress in recent years, giving rise to a range of algorithms broadly categorized into \textit{kernel-based} and \textit{matching-based} approaches. Among these, \textit{matching-based} methods, especially those based on trajectory matching, have shown excellent performance. These approaches synthesize data such that the model's training trajectory on the synthetic data closely matches the trajectory observed when training on real data, thereby enabling comparable downstream performance.

The Matching Training Trajectories \cite{cazenavette2022distillation}(MTT) was first proposed in 2022, and has since been refined by the Trajectory Matching with Soft Label Assignment \cite{10.5555/3618408.3618670}(TESLA), The Flat Trajectory Distillation \cite{DuJTZ023}(FTD) and Difficulty-Aligned Trajectory Matching \cite{DBLP:conf/iclr/Guo0CLZ024}(DATM) methods, respectively. Among them, DATM \cite{DBLP:conf/iclr/Guo0CLZ024} currently achieves state-of-the-art(SOTA) results by aligning training trajectory sub-sequences with task difficulty: early-stage trajectory parameters are preferred for simple tasks like CIFAR-10 with Image Per Class(IPC)=1, while later-stage parameters are more effective for challenging tasks like TinyImageNet with IPC=50. Despite DATM's strong performance, there remains a considerable performance gap between synthetic and real data in low-IPC settings.

To address this limitation, the goal of this study is to improve the information richness of small-scale synthetic datasets. We propose a novel dataset distillation method called DATM-CLR, which integrates contrastive learning \cite{10.5555/3524938.3525087} into the image synthesis process. Each synthetic image is augmented multiple times to create positive pairs, while samples from other categories serve as negative pairs. A projection head is applied to extract contrastive features, and a SimCLR-style contrastive loss is computed. Although methods that combine dataset distillation with contrast learning, such as the combination of dataset condensation and contrast learning \cite{pmlr-v162-lee22b}, or Mutual Information Maximization for Datset Distillation(MIM4DD) \cite{shang2023mimdd}, a method that differs from the present method in terms of contrast design, have appeared before, the present method demonstrates performance that exceeds both methods.

To further enhance the generalization and discriminative quality of the synthetic data, we design a joint optimization framework that combines trajectory matching loss and contrastive loss. This multi-task formulation ensures that the synthesized data remains both semantically consistent and feature-diverse.

%貢献とまとめ%
The contributions of this study are summarized as follows:
\begin{enumerate}
    \item We propose DATM-CLR, a new dataset distillation framework that innovatively incorporates SimCLR-style contrastive learning \cite{10.5555/3524938.3525087} into trajectory matching. This combination introduces a new paradigm for enhancing the discriminative power of synthetic data under extremely low-data regimes.
    \item Unlike prior work, we embed contrastive learning both into the synthetic image generation process and into the student model’s training updates, allowing contrastive signals to directly guide feature learning during optimization.
    \item We validate our method under low-data regimes, achieving a \textbf{6.1\%} accuracy improvement over the previous state-of-the-art DATM \cite{DBLP:conf/iclr/Guo0CLZ024} on CIFAR-10 with IPC=1, with additional gains of \textbf{0.8\%} on CIFAR-100 with IPC=1 and \textbf{1.2\%} on Tiny-ImageNet with IPC=1.
\end{enumerate}

\section{Preliminary}
\label{sec:background}

% Introduction to each method of trajectory matching %

\subsection{An Overview of Dataset Distillation}
\label{subsec:formulation_dataset_distillation}
Dataset distillation aims to synthesize a small set of informative data that allows models to achieve comparable performance to training on the full dataset \cite{wang2020datasetdistillation}. It is typically formulated as a bi-level optimization problem, where the synthetic data is optimized so that a model trained on it generalizes well on real validation data \cite{cazenavette2022distillation}. Existing methods differ mainly in how they define this optimization objective: some methods focus on matching global statistics or feature similarities between synthetic and real data \cite{DBLP:journals/corr/abs-2006-05929,50025}, while others attempt to simulate the actual training dynamics \cite{cazenavette2022distillation}. The former leads to \textit{kernel-based} approaches, and the latter gives rise to \textit{matching-based} approaches, which we detail in Sections \ref{subsec:gradient_matching} and \ref{subsec:matching_training_trajectories} respectively.

\subsection{Kernel-based Approaches}
\label{subsec:gradient_matching}
Kernel-based methods optimize synthetic data to preserve global similarity structures, often through Neural Tangent Kernels \cite{Lee_2020}(NTK) or random feature approximations. KIP(Kernel Inducing Points) formulates distillation as a kernel regression problem, finding synthetic images whose NTK-induced outputs mimic those of real data \cite{10.5555/3600270.3600983}. Random Feature Approximation for Dataset Distillation \cite{50025}(RFAD) further improves efficiency using low-rank approximations. While effective in small-scale or theoretical settings, kernel-based methods often struggle with deep networks and large-scale data, motivating dynamic matching-based alternatives.

\subsection{Matching-based Approaches}
\label{subsec:matching_training_trajectories}
Matching-based methods optimize synthetic data by directly simulating the training behavior of models on real data. Common matching objectives include gradient matching, which aligns model gradients between real and synthetic batches \cite{DBLP:journals/corr/abs-2006-05929,zhao2021datasetcondensationdifferentiablesiamese}; distribution matching, which minimizes divergence in feature or output distributions \cite{690345d30c9f4e6dbeef4ad9210bee62,9879629,10204664}; and trajectory matching \cite{cazenavette2022distillation,10.5555/3618408.3618670,DuJTZ023,DBLP:conf/iclr/Guo0CLZ024}, which aligns the parameter evolution path during training.

Among these, trajectory matching has gained increasing attention due to its ability to capture long-term training dynamics. MTT \cite{cazenavette2022distillation} first proposed to match multi-step model parameter trajectories between real and synthetic training. FTD \cite{DuJTZ023} improved MTT by introducing flat trajectory constraints to reduce accumulated error sensitivity to initialization. TESLA \cite{10.5555/3618408.3618670} further enhanced scalability on large datasets by designing a memory-efficient trajectory matching framework and introducing soft label assignment to improve performance on long-tail classes.

Building on these ideas, DATM \cite{DBLP:conf/iclr/Guo0CLZ024} proposes a dynamic trajectory selection strategy based on data difficulty: for low-IPC settings, it matches early training stages to emphasize simple patterns; for higher IPCs, it leverages later trajectories to encode complex features. In addition, DATM \cite{DBLP:conf/iclr/Guo0CLZ024} introduces soft label initialization using pretrained logits and continuously optimizes them during distillation, significantly improving generalization. DATM \cite{DBLP:conf/iclr/Guo0CLZ024} achieves state-of-the-art performance across several benchmarks and is the first to demonstrate near-lossless dataset distillation under strong compression settings.

\subsection{Contrastive Learning}
\label{subsec:contrastive_learning}
%対照学習の紹介
Contrastive learning \cite{10.5555/3524938.3525087,jaiswal2021survey,10.5555/3495724.3497291} is a self-supervised learning method, the core idea is to effectively learn semantic features from data and improve the generalization ability of the model by reducing the distance between similar samples (positive samples) and increasing the distance between dissimilar samples (negative samples). In specific implementations, contrastive learning usually designs appropriate data augmentation methods as well as loss functions \cite{10.5555/3524938.3525087,jaiswal2021survey,He2019MomentumCF}, such as InfoNCE \cite{journals/corr/abs-1807-03748} loss, to force the model to learn more robust and more discriminative representations. In recent years, contrastive learning has achieved significant success in computer vision, especially in low-resource labeling scenarios, where it outperforms traditional supervised learning \cite{10.5555/3524938.3525087,7780459,DBLP:conf/iclr/HjelmFLGBTB19}.

\section{Approach}
\label{sec:method}

\begin{figure}[h]
    \centering
    \includegraphics[width=\linewidth]{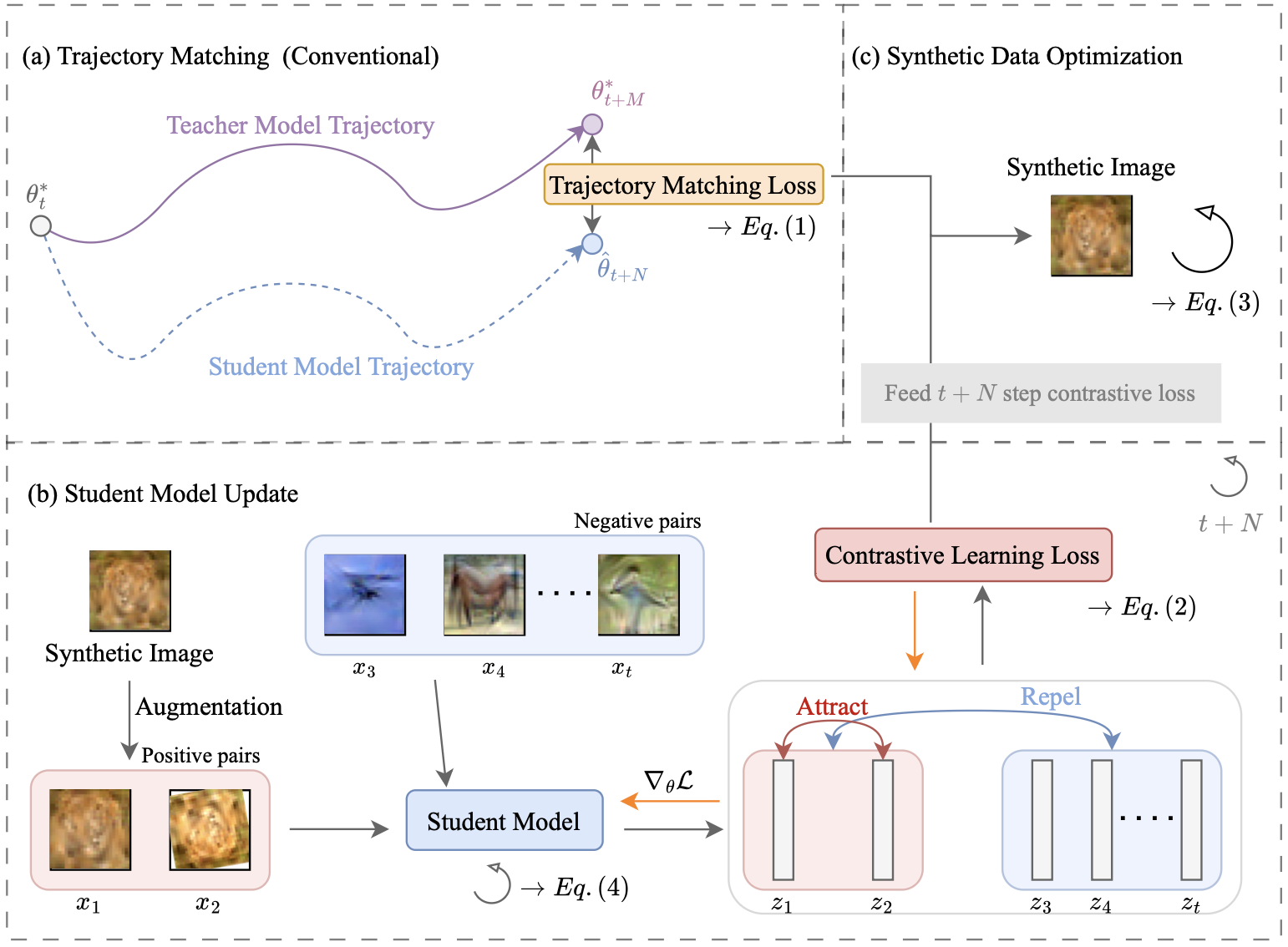}
    \caption{Illustration of the synthetic data optimization process integrating Trajectory Matching and Contrastive Learning. (a) Trajectory Matching aligns the student model parameter trajectory trained on synthetic data with the teacher model parameter trajectory obtained from real data. (b) Contrastive Learning generates positive pairs via augmentation from synthetic images and distinguishes them from negative pairs in the feature space, encouraging intra-class similarity and inter-class dissimilarity. In addition, the Contrastive Update strategy leverages the contrastive loss to directly update the student model during its inner-loop optimization, enabling synthetic data to better capture discriminative representations. (c) Synthetic images are optimized simultaneously by minimizing trajectory matching loss and contrastive learning loss.}
    \label{fig:method}
\end{figure}

In this section, we introduce our proposed method, Difficulty-Aligned Trajectory Matching with Simple Framework for Contrastive Learning of Visual Representations (DATM-CLR). We first review the basic idea of Trajectory Matching  \cite{cazenavette2022distillation}(TM) and then present the specific design and implementation details of our proposed method in Fig~\ref{fig:method}.

\subsection{Problem Definition}
\label{subsec:problem difinition}
Dataset Distillation (DD) aims to generate a much smaller synthetic dataset $\mathcal{D}_{\text{syn}}$ from a large real dataset $\mathcal{D}_{\text{real}}$, such that a model trained on $\mathcal{D}_{\text{syn}}$ achieves comparable performance to one trained on $\mathcal{D}_{\text{real}}$. Trajectory Matching (TM) is one of the most advanced methods in dataset distillation. It optimizes the synthetic dataset to make the training trajectory of a student model on syntetic dataset closely match that on the real dataset, thereby effectively simulating long-term training dynamics.

Specifically, given the parameter trajectory $\{\theta_t^*\}_{t=0}^T$ obtained by training the teacher model on the real dataset, trajectory matching seeks to optimize the synthetic dataset such that the student model, initialized at step $t$, reaches a parameter state close to the teacher’s state at step $t+M$ after $N$ training steps:

\begin{equation}
\mathcal{L}_{\text{tm}} = \frac{\left\| \hat{\theta}_{t+N} - \theta^{*}_{t+M} \right\|_2^2}{\left\| \theta^{*}_{t} - \theta^{*}_{t+M} \right\|_2^2}
\end{equation}

Here, $\hat{\theta}_{t+N}$ denotes the model parameters obtained after $N$ steps of training on the synthetic dataset, and $\theta_{t+M}^{*}$ is the corresponding target parameter from the teacher’s trajectory. The denominator serves as a normalization term representing the actual change in the teacher’s parameters over $M$ steps.

\subsection{Incorporating Contrastive Learning}
\label{subsec:incorporating contrastie learning}
First, we adapt the basic framework of the DATM \cite{DBLP:conf/iclr/Guo0CLZ024} method. By matching the training trajectory, we dynamically adjust the learning stages of the synthetic dataset to adapt its difficulty to that of the real dataset. Specifically, we select the matching point based on the size of the synthetic dataset and the dynamic trajectory of the model, and use the soft labels of the real dataset to improve the expression quality of the synthetic data.

Next, we incorporate contrastive learning to the trajectory matching. Inspired by SimCLR \cite{10.5555/3524938.3525087}, we apply the contrastive loss $\mathcal{L}_{\text{contrast}}$ to encourage the model to learn discriminative and robust feature representations, which can effectively distinguish between similar and dissimilar samples. In our proposed method, we select two groups of synthetic data with different augmentations from the current synthetic dataset in every training iteration. Then using class labels to construct positive and negative pairs. It is important to emphasize that, unlike conventional contrastive learning, which typically assumes unlabeled data, our approach explicitly leverages label information, thus constituting supervised contrastive learning \cite{10.5555/3495724.3497291}. A projection head is applied to map the feature embeddings to a lower-dimensional space, followed by contrastive loss calculation based on normalized temperature-scaled cross-entropy, enabling the synthetic samples to be distinguishable and semantically meaningful.

Specifically, we adapt the contrastive loss form of SimCLR \cite{10.5555/3524938.3525087} as follows:

\begin{equation}
\mathcal{L}_{\text{contrast}} = -\log \frac{\exp(\text{sim}(z_1, z_2)/\tau)}{\exp(\text{sim}(z_1, z_2)/\tau) + \sum_{i=3}^N \exp(\text{sim}(z_1, z_i)/\tau)},
\end{equation}

\noindent
where $\text{sim}(\cdot, \cdot)$ is the cosine similarity function, $\tau$ is the temperature parameter controlling the difficulty of contrastive learning, and $z_1$, $z_2$, $z^-$ are the positive and negative feature representations obtained from augmented synthetic samples.

We combine this contrastive loss with the trajectory matching loss $\mathcal{L}_{\text{tm}}$ to jointly guide the optimization of the synthetic dataset. This hybrid objective allows the synthetic data to not only match the real trajectory but also contain discriminative semantic features, thereby improving the quality and generalizability of the synthetic data.

Finally, the combined loss for updating synthetic data is as follows:

\begin{equation}
\mathcal{L}_{\text{total}} = \alpha \mathcal{L}_{\text{contrast}}(z_1, z_2, z^-) +  \beta \mathcal{L}_{\text{tm}}(\hat{\theta}_{t+N}, \theta^*_{t+M}),
\end{equation}

\noindent
where $\alpha, \beta$ are tunable loss weights; $\hat{\theta}_{t+N}$ denotes the final model parameter after $t+N$ inner updates on the student network; $\theta^*_{t+M}$ is the target teacher parameter at step $t+M$; $\mathcal{L}_{\text{contrast}}$ encourages the synthetic dataset to form a well-clustered feature space; and $\mathcal{L}_{\text{tm}}$ ensures that the student's parameter trajectory matches that of the teacher. We call this update strategy '\textbf{Contrastive Fusion}'.

Our experimental results show that this strategy has clear advantages. Unlike other methods that rely solely on parameter matching, our method captures finer-grained features in fewer update steps. Although reducing the size of the synthetic dataset typically results in a performance drop, our approach preserves strong generalization by combining label-aware trajectory matching with contrastive objectives. By jointly optimizing the training trajectory and feature semantics, the synthetic data produced by our method can effectively simulate real data learning behavior, enabling us to achieve performance close to real-data-trained models.
Therefore, we belive this method as the main realization path of our approach to obtain high-quality, compact synthetic datasets.

Moreover, in addition to this strategy of Puls, we propose another strategy that makes a combination of contrast loss and trajectory matching loss. When updating the student network parameters via gradient descent, we jointly minimize cross-entropy loss $\mathcal{L}_{\text{ce}}$ and contrastive loss $\mathcal{L}_{\text{contrast}}$, which we call the '\textbf{Contrastive Update}' strategy, as follows:

\begin{equation}
\hat{\theta}_{n+1} = \hat{\theta}_n - \alpha_{\text{syn}} \nabla_{\hat{\theta}_n} \left( \mathcal{L}_{\text{ce}} \left( f_{\hat{\theta}_n}(x), y \right) + \lambda \mathcal{L}_{\text{contrast}}(z_1, z_2, z^-) \right)
\end{equation}

\noindent
where $\alpha_{\text{syn}}$ is the trainable learning rate, $\lambda$ is the weight of the contrastive loss, $f_{\hat{\theta}_n}$ denotes the current student network, and $(x, y)$ is a mini-batch sampled from the synthetic dataset.

By directly applying contrastive learning during the gradient update process, we update the student network parameters in each iteration while simultaneously computing both contrastive loss and parameter matching loss. This strategy allows model parameters to be updated while preserving class-level distinctions in the synthetic data, which helps the model capture fine-grained features in the synthetic samples.For results, please refer to Section~\ref{ablation}

\subsection{Algorithmic Flow}
\label{subsec:algorithmice flow}
To elaborate on $Algorithm 1$, our method builds upon the DATM framework \cite{DBLP:conf/iclr/Guo0CLZ024} by introducing contrastive learning \cite{10.5555/3524938.3525087} during the inner optimization loop of dataset distillation. The motivation is to encourage the synthetic data to capture not only training dynamics (via trajectory matching) but also discriminative features across classes (via contrastive learning).

%手法のアルゴリズム%
\begin{algorithm}[H]
\caption{Dataset Distillation Enhanced by Contrastive Learning}
\begin{algorithmic}[1]
\Require Teacher trajectories $\{\theta_t^*\}$, augmentation $\mathcal{A}$, inner steps $N$, learning rates $\alpha_{\text{syn}}$, loss weights $\alpha$, $\beta$
\State Initialize synthetic data $\mathcal{D}_{\text{syn}}$ and soft labels
\For{each distillation iteration}
    \State Sample expert trajectory and starting point $t$
    \State Set student weights $\hat{\theta}_0 \gets \theta_t^*$
    \For{$n = 0$ to $N-1$}
        \State Sample batch $(x, y)$ from $\mathcal{D}_{\text{syn}}$, apply $\mathcal{A}$: $(x_1, x_2)$
        \State Sample negatives $x^-$ from other classes
        \State Encode features $z_1, z_2, z^-$ using projector
        \State Compute contrastive loss $\mathcal{L}_{\text{contrast}}(z_1, z_2, z^-)$
        \State Compute CE loss $\mathcal{L}_{\text{ce}} = \ell(f_{\hat{\theta}_n}(x_1), y)$
        \State Update student: $\hat{\theta}_{n+1} \gets \hat{\theta}_n - \alpha_{\text{syn}} \nabla(\mathcal{L}_{\text{ce}})$
    \EndFor
    \State Compute trajectory loss $\mathcal{L}_{\text{tm}} = \frac{\left\| \hat{\theta}_{t+N} - \theta^{*}_{t+M} \right\|_2^2}{\left\| \theta^{*}_{t} - \theta^{*}_{t+M} \right\|_2^2}$
    \State Total loss: $\mathcal{L} = \alpha \cdot \mathcal{L}_{\text{contrast}} + \beta \cdot \mathcal{L}_{\text{tm}}$
    \State Update $\mathcal{D}_{\text{syn}}$, $\alpha_{\text{syn}}$, soft labels via gradient descent
\EndFor
\Ensure Optimized synthetic dataset $\mathcal{D}_{\text{syn}}$
\end{algorithmic}
\end{algorithm}

The algorithm starts by initializing the synthetic dataset $\mathcal{D}{\text{syn}}$ and corresponding soft labels. In each distillation iteration, a teacher trajectory segment is selected, and a student model is initialized with the corresponding weights. Then, for a fixed number of inner steps, batches from $\mathcal{D}{\text{syn}}$ are augmented twice to form positive pairs $(x_1, x_2)$, while negative samples $x^-$ are drawn from different classes. A projection head encodes these samples into latent features, which are then used to compute a SimCLR-style contrastive loss. At the same time, a cross-entropy loss is computed using one of the augmented views and the soft label.

Only the cross-entropy loss is used to update the student model, preserving compatibility with the trajectory matching framework \cite{cazenavette2022distillation}. After the student model reaches the end of the simulated training steps, we compute a trajectory matching loss that measures the distance between the final student parameters and the teacher parameters. This loss is normalized to improve stability across different tasks and scales.

Finally, a joint loss function combining the contrastive loss and trajectory loss is used to update the synthetic data, soft labels, and learnable learning rate via gradient descent. This joint optimization allows the synthetic data to better align with both the feature distribution and the learning dynamics of the real dataset.

This design effectively integrates semantic supervision into the trajectory-based optimization, leading to synthetic data that is more informative and transferable, especially in low-IPC settings.

\section{Experiments}
\label{sec:experiments}

\subsection{Settings}
\label{subsec:experiment_setting}
We conduct experiments on the CIFAR-10, CIFAR-100 \cite{krizhevsky2009learning}, and Tiny-ImageNet datasets \cite{Le2015TinyIV} following the standard protocols. We evaluate several CNN architectures (ConvNet \cite{zhao2021datasetcondensationdifferentiablesiamese,cazenavette2022distillation,loo2023datasetdistillationconvexifiedimplicit}, ResNet18 \cite{7780459}, AlexNet \cite{10.1145/3065386}, VGG \cite{journals/corr/SimonyanZ14a}, DenseNet \cite{DBLP:conf/cvpr/HuangLMW17}) to validate the generalizability of the distilled datasets.

In our DATM-CLR method, we employ Differentiable Siamese Augmentation \cite{zhao2021datasetcondensationdifferentiablesiamese} (DSA) techniques during the distillation process, including random cropping, flipping, color jittering, and affine transformations. Specifically, we jointly optimize the synthetic data $\mathcal{D}_{\text{syn}}$, soft labels, and learning rates $\alpha_{\text{syn}}$ via gradient descent by minimizing a combined trajectory matching loss $\mathcal{L}_{\text{tm}}$ and a contrastive loss $\mathcal{L}_{\text{contrast}}$. At each distillation iteration, student parameters are iteratively updated for $N=20$ inner steps using SGD with momentum $0.5$. Subsequently, the trajectory matching loss $\mathcal{L}_{\text{tm}}$ is calculated between the student parameters after $N$ updates and the corresponding teacher parameters at $M$ future steps. Additionally, the SimCLR \cite{10.5555/3524938.3525087} contrastive temperature is set to $0.1$, and random seeds are fixed to ensure reproducibility.

For evaluation, we randomly initialize the weights of multiple neural networks on the distilled dataset and report the average accuracy over five trials with different random seeds on the original test set. The experiments were conducted using single NVIDIA GeForce RTX3090 (24GB) and an Intel Core i9-10850K 3.60GHz (32GB).

\subsection{Efficiency of DATM-CLR}
\label{subsec:results}

\begin{table}[ht]
\centering
\begin{tabular}{l|ccc|cc|c}
\hline
\textbf{Dataset} & \multicolumn{3}{c|}{\textbf{CIFAR-10}} & \multicolumn{2}{c|}{\textbf{CIFAR-100}} & \textbf{Tiny-ImageNet} \\
\textbf{IPC}     & 1            & 10           & 50           & 1             & 10            & 1              \\ \hline
Random           & 15.4±0.3     & 31.0±0.5     & 50.6±0.3     & 4.2±0.3       & 14.6±0.5      & 1.4±0.1        \\
MTT              & 46.2±0.8     & 65.4±0.7     & 71.6±0.2     & 24.3±0.3      & 39.7±0.4      & 8.8±0.3        \\
TESLA            & 48.5±0.8     & 66.4±0.8     & 72.6±0.7     & 24.8±0.4      & 41.7±0.3      & --             \\
FTD              & 46.0±0.4     & 65.3±0.4     & 73.2±0.2     & 24.4±0.4      & 42.5±0.2      & 10.5±0.2       \\
DATM             & 46.9±0.5     & 66.8±0.2     & 76.1±0.3     & 27.9±0.2      & 47.2±0.4      & 13.6±0.3       \\
\textbf{Ours}    & \textbf{53.0±0.4} & \textbf{70.2±0.3} & \textbf{77.6±0.3} & \textbf{28.7±0.2} & \textbf{48.7±0.3} & \textbf{14.8±0.2} \\ \hline
Full Dataset     & \multicolumn{3}{c|}{84.8±0.1}     & \multicolumn{2}{c|}{56.2±0.3}    & 37.6±0.4       \\ \hline
\end{tabular}
\caption{Test accuracy comparison across different dataset distillation methods on CIFAR-10, CIFAR-100, and Tiny-ImageNet with varying Images Per Class (IPC). The proposed method ("Ours") consistently outperforms prior methods and achieves the closest performance to full dataset training, especially under low-data regimes.}
\label{tab:distillation_comparison}
\end{table}

\begin{wraptable}{r}{0.5\textwidth}
\small
\begin{tabular}{c|c|c|c|c}
\hline
Method & ConvNet        & ResNet18       & VGG            & AlexNet \\ \hline
Random & 15.46          & 13.97          & 12.07          & 10.55 \\ \hline
DATM   & 46.91          & 44.70          & 43.62          & 43.68\\ \hline
Ours   & \textbf{53.00} & \textbf{49.36} & \textbf{48.98} & \textbf{48.55} \\ \hline
\end{tabular}
\caption{Generalization performance of distilled datasets across architectures. The proposed method ("Ours") consistently outperforms Random and DATM on all networks, demonstrating strong generalization.}
\end{wraptable}

We compare the performance of our method with existing dataset distillation approaches on CIFAR-10, CIFAR-100, and Tiny-ImageNet. The results demonstrate that our proposed method, which integrates contrastive learning into DATM, consistently achieves state-of-the-art performance across all datasets and IPC settings.

First, on the \textbf{CIFAR-10}, under the IPC=1 setting (i.e., one synthetic image per class), our method achieves an accuracy of 53.0\%, outperforming the best baseline DATM method (46.9\%) by \textbf{6.1\%}, demonstrating superior performance in extreme low-data regimes. Furthermore, as IPC increases to 50, our method still achieves an accuracy of 70.2$\pm$0.3\%, compared to 66.8$\pm$0.3\%, consistently outperforming existing methods including DATM\cite{DBLP:conf/iclr/Guo0CLZ024}, FTD\cite{DuJTZ023}, and TESLA\cite{10.5555/3618408.3618670}.

Second, on the more challenging \textbf{CIFAR-100}, our method shows significant advantages under low IPC. At IPC=1, our method achieves an accuracy of 28.7\%, outperforming the best baseline DATM method (27.9\%) by \textbf{0.8\%}. Although the improvement is smaller compared to CIFAR-10, it remains statistically significant. At IPC=50, our method reaches 48.7$\pm$0.3\%, outperforming both DATM (47.2$\pm$0.4\%) and other mainstream methods.

Lastly, on the large-scale and highly-difficulty \textbf{Tiny-ImageNet}, our method achieves 14.8\% accuracy under IPC=1, outperforming the best existing method (13.6\%) by \textbf{1.2\%}. This result demonstrates that our approach maintains effectiveness even in large-scale, high-resolution, and complex scenarios.

In summary, through extensive experiments, we show that our distillation method effectively distills high-quality, transferable synthetic datasets. It demonstrates superior performance, especially under low IPC settings. This confirms that incorporating contrastive learning improves the semantic clustering of synthetic data, allowing it to better approximate the distribution of real data and thereby enhance training dynamics and final performance.

\subsection{Ablation}
\label{ablation}

In the optimization process of synthetic data, we not only consider the combination method of trajectory matching loss and contrast loss, but also consider how different data enhancement will bring about changes. Specifically, it is shown as follows:

\begin{itemize}
    \item the Contrastive Fusion and Contrastive Update strategies proposed in Section~\ref{subsec:incorporating contrastie learning} based on different combining methods.
    \item the distinction between using single and multiple methods of data augmentation
\end{itemize}

In addition to this, we performed ablation experiments on the hyperparameters in the Contrastive Fusion and Contrastive Update strategies in Tab \ref{tab:ablation}.

\begin{table}[h]
\renewcommand{\arraystretch}{1.3}
\centering
\begin{tabular}{l@{\hspace{10pt}}c@{\hspace{10pt}}c|c@{\hspace{10pt}}c|c@{\hspace{10pt}}c}
\toprule
\textbf{Strategy} & \textbf{Single} & \textbf{Multiple} & \textbf{Single} & \textbf{Multiple} & \textbf{Single} & \textbf{Multiple} \\
\midrule
\multicolumn{3}{c|}{CIFAR-10 IPC1} & 
\multicolumn{2}{c|}{CIFAR-10 IPC10} & 
\multicolumn{2}{c}{CIFAR-100 IPC1} \\
\midrule
Contrastive Update & 44.92 & 46.72 & 66.22 & 67.70 & 24.92 & 26.23 \\
Contrastive Fusion  & 48.25 & \textbf{53.00} & 68.24 & \textbf{70.23} & 27.60 & \textbf{28.74} \\
\bottomrule
\end{tabular}
\caption{Comparison of two strategies for integrating contrastive learning into the trajectory matching framework. "Single" applies contrastive loss within each inner-loop update step alongside cross-entropy loss, promoting feature discriminability of synthetic samples during every gradient update. "Multiple" denotes an alternative strategy using multiple synthetic trajectories. Results under various datasets and settings demonstrate that the "Contrastive Fusion" method consistently improves training stability and data quality.}
\label{tab:ablation}
\end{table}

\begin{figure}[h]
    \centering
    \begin{subfigure}[b]{0.45\textwidth}
        \centering
        \includegraphics[width=\textwidth]{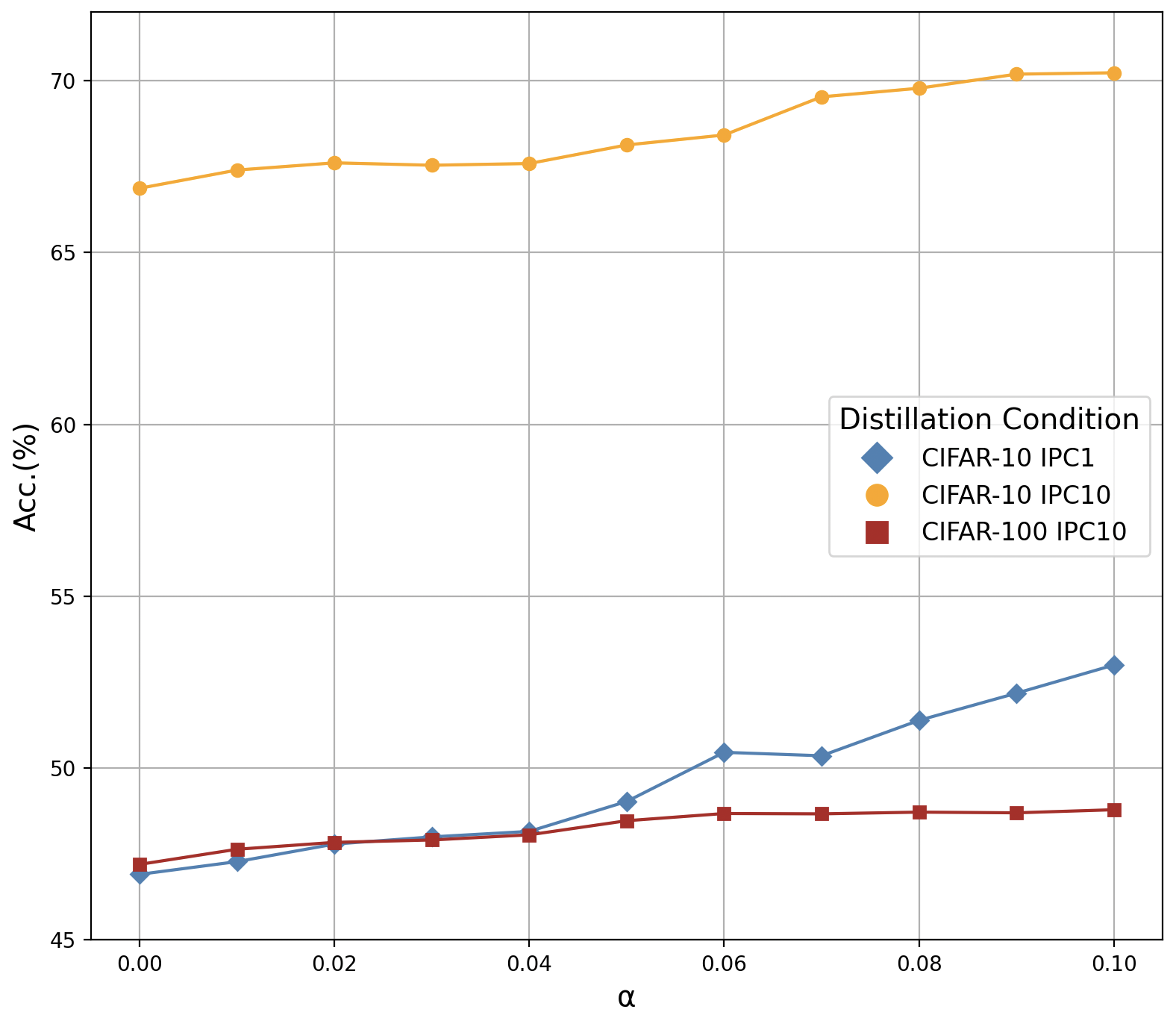}
        \caption{In 'Contrastive Fusion' strategy}
        \label{fig:ablation1}
    \end{subfigure}
    \hfill
    \begin{subfigure}[b]{0.45\textwidth}
        \centering
        \includegraphics[width=\textwidth]{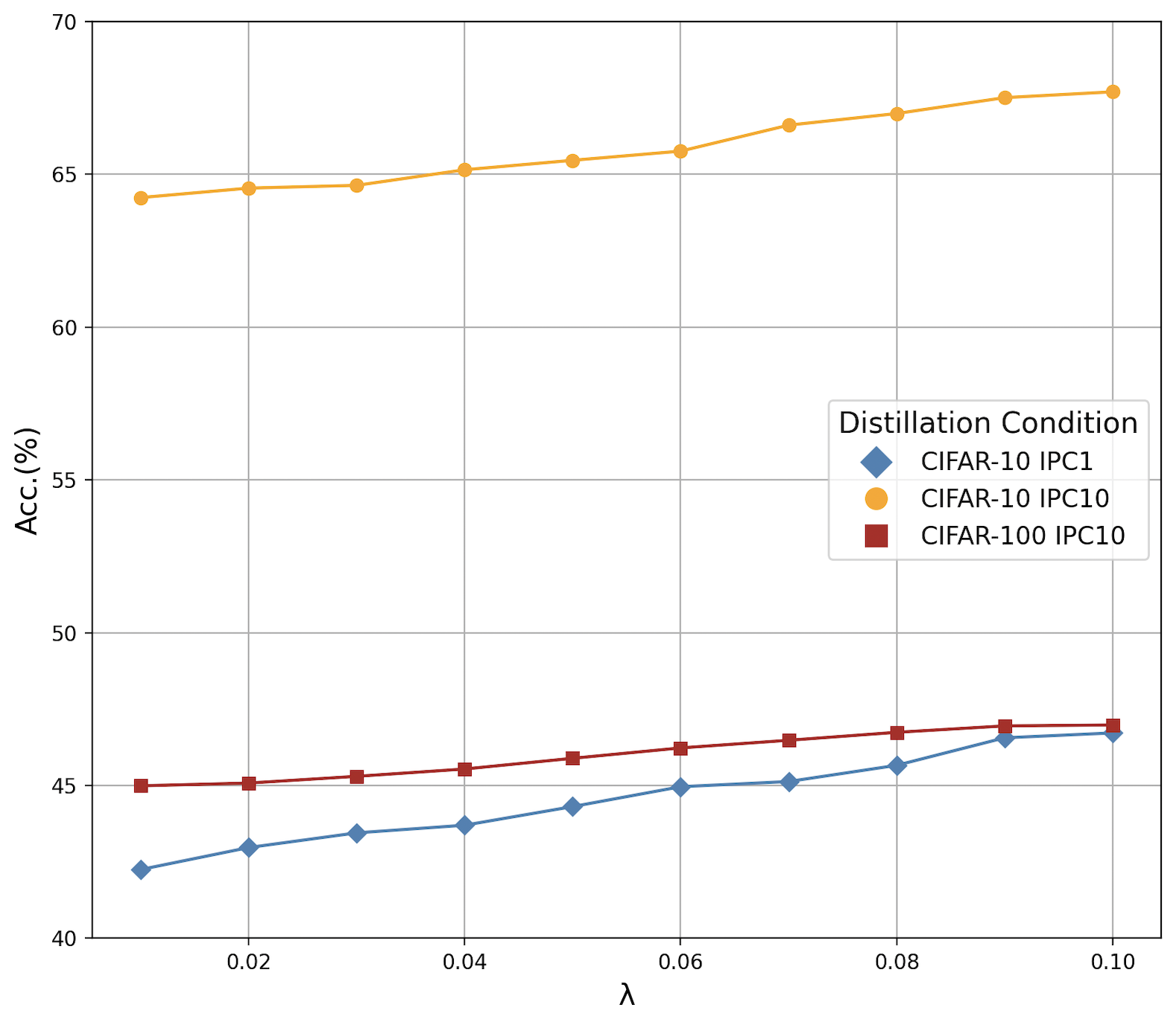}
        \caption{In 'Contrastive Update' strategy}
        \label{fig:ablation2}
    \end{subfigure}
    \caption{Variation of different effects of hyperparameters in different strategies}
    \label{fig:ablation}
\end{figure}

In order to investigate the impact of contrasting learning losses with different emphasis in our loss formulation, we conducted an extensive ablation study. We fixed $\lambda$ to 1, and conducted extensive ablation studies by systematically varying the hyperparameter $\alpha$.
As shown in Fig \ref{fig:ablation1}, the performance of the model varies significantly with different values of $\alpha$. The picture shows that the model accuracy shows a steady upward trend with increasing $\alpha$ from 0 to 0.1. In summary, increasing the weight of contrast loss helps to enhance the discriminative ability of the synthetic data, which significantly improves the model distillation performance.

We also performed ablation experiments on the hyperparameter $\lambda$ in the Contrastive Update strategy in three settings as shown in Fig \ref{fig:ablation2}. The experimental results show that the performance of the distillation dataset is decreasing with the increase of $\lambda$ in all the settings, indicating that too much weight of the regular term will weaken the optimization effect of the main objective and affect the performance of the distillation dataset. This trend suggests that $\lambda$ needs to be parameterized in practical applications, where $\lambda$ = 0.1 tends to achieve better performance in different tasks.

\section{Conclusion}
This paper addresses the critical challenge of performance degradation in dataset distillation under extremely low-data regimes. By proposing DATM-CLR, a method that integrates SimCLR-style contrastive learning into the DATM trajectory matching framework, we aim to enhance the semantic richness and discriminative capability of synthetic datasets. Our approach directly tackles the objective introduced at the beginning: narrowing the performance gap between real and synthetic data when the number of images per class (IPC) is extremely low.

Through a joint optimization of trajectory matching loss and contrastive loss, our method enables synthetic samples to not only align with the training dynamics of real data but also capture stronger class-wise semantic structures. This dual optimization significantly improves both the optimization quality and the generalization performance of distilled datasets.

Extensive experiments on CIFAR-10, CIFAR-100, and Tiny-ImageNet confirm the effectiveness of our method. In particular, on CIFAR-10 with IPC=1, our method outperforms the current state-of-the-art DATM by $6.1\%$, demonstrating the potential of combining trajectory and contrastive learning.

Despite its effectiveness, our approach introduces additional limitations. The integration of contrastive learning increases the computational burden and introduces more hyperparameters, which may complicate deployment in resource-constrained settings. Furthermore, tuning the balance between contrastive and trajectory loss remains non-trivial and dataset-dependent.

Nevertheless, our work offers new insights into dataset distillation. It provides a promising direction for future research—how to jointly optimize in both parameter space and semantic feature space to generate high-quality, compact, and semantically meaningful synthetic datasets.

\begin{credits}
\subsubsection*{\ackname} This work was supported in part by the Japan Society for the Promotion of Science (JSPS) KAKENHI Grant-in-Aid for Scientific Research (C) under Grant 23K11164. This work was also supported by several competitive funds within the University of Fukui.
\end{credits}

% \subsubsection{\discintname}
% It is now necessary to declare any competing interests or to specifically
% state that the authors have no competing interests. Please place the
% statement with a bold run-in heading in small font size beneath the
% (optional) acknowledgments\footnote{If EquinOCS, our proceedings submission
% system, is used, then the disclaimer can be provided directly in the system.},
% for example: The authors have no competing interests to declare that are
% relevant to the content of this article. Or: Author A has received research
% grants from Company W. Author B has received a speaker honorarium from
% Company X and owns stock in Company Y. Author C is a member of committee Z.

\bibliographystyle{splncs04}
\bibliography{mybib}

\end{document}